\documentclass[nohyperref]{article}

\usepackage{microtype}
\usepackage{graphicx}
\usepackage{subfigure}
\usepackage{booktabs} 

\usepackage{hyperref}

\usepackage[accepted]{ai4science}

\usepackage{amsmath}
\usepackage{amssymb}
\usepackage{pifont}
\usepackage{mathtools}
\usepackage{amsthm}
\usepackage{xcolor}

\usepackage{booktabs}
\usepackage{multirow}
\usepackage{comment}
\usepackage{graphicx}
\usepackage{makecell}
\usepackage{siunitx}
\usepackage{booktabs}

\usepackage[capitalize,noabbrev]{cleveref}

\theoremstyle{plain}

\theoremstyle{definition}

\theoremstyle{remark}

\newcommand{\cmark}{\ding{51}}%
\newcommand{\xmark}{\ding{53}}%

\begin{document}

\twocolumn[
\icmltitle{Learning the Solution Operator of Boundary Value Problems using Graph Neural Networks}



\icmlsetsymbol{equal}{*}
\icmlsetsymbol{doneAtMxm}{$\dagger$}

\begin{icmlauthorlist}
\icmlauthor{Winfried L\"otzsch}{MXM}
\icmlauthor{Simon Ohler}{MXM,TUK,doneAtMxm}
\icmlauthor{Johannes S. Otterbach}{MXM}
\end{icmlauthorlist}

\icmlaffiliation{MXM}{Merantix Momentum, AI Campus Berlin, Germany}
\icmlaffiliation{TUK}{Univ. of Kaiserslautern, Kaiserslautern, Germany}

\icmlcorrespondingauthor{Winfried L\"otzsch}{Winfried.loetzsch@merantix.com}
\icmlcorrespondingauthor{Johannes Otterbach}{johannes.otterbach@merantix.com}

\icmlkeywords{Graph Neural Networks, AI4Science}

\vskip 0.3in
]

\printAffiliationsAndNotice{\workDoneAtMxm} 


\begin{abstract}
   As an alternative to classical numerical solvers for partial differential equations (PDEs) subject to boundary value constraints, there has been a surge of interest in investigating neural networks that can solve such problems efficiently. In this work, we design a general solution operator for two different time-independent PDEs using graph neural networks (GNNs) and spectral graph convolutions. We train the networks on simulated data from a finite elements solver on a variety of shapes and inhomogeneities. In contrast to previous works, we focus on the ability of the trained operator to generalize to previously unseen scenarios. Specifically, we test generalization to meshes with different shapes and superposition of solutions for a different number of inhomogeneities. We find that training on a diverse dataset with lots of variation in the finite element meshes is a key ingredient for achieving good generalization results in all cases. With this, we believe that GNNs can be used to learn solution operators that generalize over a range of properties and produce solutions much faster than a generic solver. Our dataset, which we make publicly available, can be used and extended to verify the robustness of these models under varying conditions.
\end{abstract}

\section{Introduction}
\begin{figure}
    \centering
    \includegraphics[trim={0 7cm 0cm 0cm},clip,width=\linewidth]{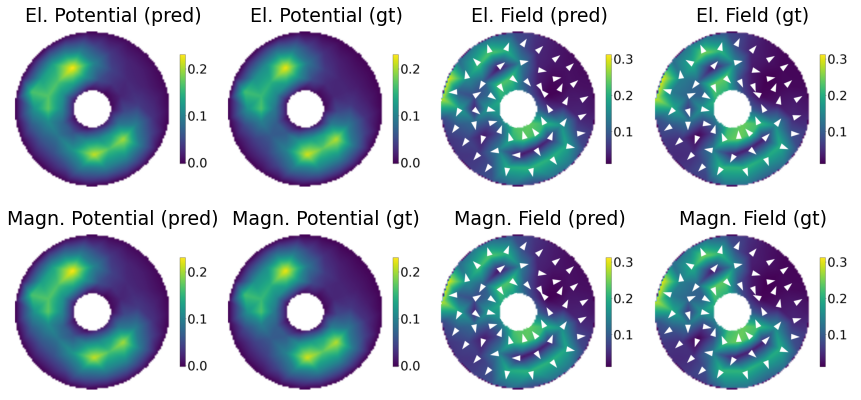}
    \vspace{-5mm}
    \caption{We train a neural network to predict solutions of boundary value problems and additional quantities like in this case the electric potential and electric field of an electrostatics simulation. Here we compare predictions of the model (pred) with ground truth data from an FEM simulation (gt). For the electric field, we visualize the magnitude of the field and overlay it with arrows depicting the field orientation.}
    \vspace{-5mm}
    \label{fig:teaser}
\end{figure}
\label{sec:intro}
Graph Neural Networks (GNNs)~\cite{Scarselli2009TheGN} have recently seen a plethora of new applications ranging from molecule and protein property modeling~\cite{Jumper2021HighlyAP, henderson2021improving, maziarka2020molecule}, learning complex dynamics from data ~\cite{Battaglia2016InteractionNF, Pfaff2021LearningMS} to combinatorial optimization ~\cite{cappart2021combinatorial}. Many real-world applications produce data artifacts that are naturally expressed as graphs, such as knowledge databases, social networks or supply chains. 

In this paper, we focus on a particular application domain that is most naturally defined by graphs: the explicit solution of time-independent (static) boundary value problems. Solutions to such problems are only implicitly defined via partial differential equations (PDEs) subject to boundary value constraints. Solving boundary value problems is required in many physics disciplines, such as thermodynamics or electromagnetics. Finite element methods (FEM), which discretize the PDE on a lattice, are commonly used as generic solvers. The resulting matrix equation is solved using exact or iterative methods of linear algebra. While these methods produce high-quality results, they typically are numerically expensive and any variations on the parameters of the problem require a full rerun to obtain the solutions.

To address this problem, we propose to use GNNs to learn the solution operator of a class of PDEs similar to previous works~\cite{Li2020NeuralOG, Li2021FourierNO}. The triangulation of the domain as a first step of an FEM solver is faster than numerically computing the solution of the PDE, which involves an expensive matrix inversion operation. We use only the triangulation grid of an FEM solution as the input to a graph network and employ supervised graph convolution operations to learn the value of the solution function at the triangulation points. We complement previous works that use GNNs to solve time-dependent problems~\cite{Pfaff2021LearningMS, SanchezGonzalez2020LearningTS, Brandstetter2022MessagePN}. We compare an example prediction of our model to the corresponding FEM solution in figure~\ref{fig:teaser}.

To show that the network approximates the solution operator, we investigate if it can learn \textit{superpositions} of different solutions and generalize to different meshes (\textit{shape generalization}) beyond the square mesh that is used frequently in the literature so far \cite{Li2020NeuralOG, Li2021FourierNO}. 

GNNs are naturally suited to handle non-square geometries as opposed to CNNs, which are commonly defined on square lattices. CNNs can be seen as a special case of GNNs on regular square domains and a fixed number of neighbors for each node. Hence, CNNs need to work on interpolated solutions if the structure of the data deviates or non-square shapes are used. Moreover, commonly used CNNs need to use masks if parts of the mesh are defined as cut-out regions. We show that our proposed GNN naturally solves all these cases within a unified approach. To our knowledge, this is the first in-depth analysis of GNNs applied to static PDEs that goes beyond learning the solution for a single mesh. Specifically, our contributions are:
\begin{itemize}
    \item We show that a GNN can efficiently learn the solution to a PDE from simulation data without explicit access to the PDE.
    \item We demonstrate that diversity with regards to the shape of meshes is a key ingredient for obtaining robust solution operators.
    \item We demonstrate that our GNN is able to generalize to previously unseen geometries and learn superpositions. As such, we obtain an approximation of the solution operator, instead of a solution itself. 
    \item We construct a new dataset consisting of tesselations and solutions of PDEs in the domains of electrostatics and magnetostatics under non-trivial boundary conditions.
\end{itemize}

Especially the shape generalization task which we design is considerably more complex than similar tasks in the recent literature, e.g. \cite{Li2020NeuralOG} which uses only square meshes. We not only verify that the network can adapt to multiple differently-shaped meshes at the same time, but also that it generalizes to a different mesh, which has not been part of the training data. We make both the created dataset\footnote{\url{https://github.com/merantix-momentum/squirrel-datasets-core}} as well as the code for our experiments publicly available\footnote{\url{https://github.com/merantix-momentum/gnn-bvp-solver}}. 

\section{Related Work}
Battaglia et al.~\yrcite{Battaglia2016InteractionNF} introduce GNNs for the simulation of the time dynamics of physical objects under complex boundary constraints, such as a string draping over a pole under gravity. They learn the dynamics from purely observational data obtained via simulations of the physical system. They show that the learned dynamics generalize to new situations with a significant speedup over the traditional simulation.
Sanchez-Gonzalez et al.~\yrcite{SanchezGonzalez2020LearningTS} extend this approach and learn the complex dynamics governing the motion of fluids and deformable bodies from simulated data. Each particle represents a node in a graph and the network learns the time update operator from the observed data.
Pfaff et al.~\yrcite{Pfaff2021LearningMS} are expanding on this by simulating the time dynamics of complex processes based on learning from meshed data. They use two types of nodes to, e.g., model a piece of fabric in the wind held by a flag-pole. Moreover, they show that the learned dynamics generalize from 2k to 20k nodes in the graph despite the latter being out-of-distribution.
Mayr et al.~\yrcite{Mayr2021BoundaryGN} use a GNN to learn the dynamics of a granular material flowing inside a non-stationary boundary, e.g. a rotating drum, by introducing a method to update the nodes describing the geometric boundaries in the GNN.

Fang et al.~\yrcite{Fang2020APN} use physics-inspired neural networks (PINNs) to parameterize the solution to time-independent PDEs. Making use of the auto-differentiation framework, they minimize the error in the PDE-solution. They develop solutions to common PDEs on complicated two- and three-dimensional manifolds. A similar approach is developed by Cai et al.~\yrcite{Cai2022PhysicsinformedNN}, who simulate fluid dynamics by combining simulation data and physics-based priors.
Afshar et al.~\yrcite{Afshar2019PredictionOA} use a CNN architecture to predict the static flow field of a hydrodynamics problem from a triangulated version of the underlying geometry.
Yao et al.~\yrcite{Yao2020FEANetAP} leverage a PINN based on the combination of physics-based Finite Element Analysis (FEA) and CNNs to create an efficient predictor of mechanical response quantities.

Belbute-Peres et al.~\yrcite{BelbutePeres2020CombiningDP} integrate an industry-grade PDE solver as a layer into a GNN simulator. They use adjoint methods to compute the gradient of the solver efficiently. They apply this approach to the simulation of complex fluid dynamics problems.
Chamberlain et al.~\yrcite{Chamberlain2021GRANDGN} reverse the approach and leverage discretization schemes developed for numerical simulations of PDEs to design new graph aggregation operations. They show that this approach leads to superior results on some standard GNN benchmarks.

Aarts et al.~\yrcite{Aarts2004NeuralNM} use a precursor to PINNs based on fully connected layers to study the solution space of a neural network applied to simple initial as well as boundary value problems. Palade et al.~\yrcite{Palade2020NeuralNA} use neural networks with radial basis functions and an integral objective to solve elliptic PDEs with non-local boundary conditions. However, both approaches do not leverage data to solve the problem, but rather parameterize the solution to a specific instance. 

Alet et al.~\yrcite{Alet2019GraphEN} use a Graph Element Network (GEN) to learn a mapping from initial conditions to the final output of an FEM-based numerical simulation. They show that the GEN can be used to optimize the FEM mesh to focus on the non-linear part of the solution space. A similar approach is also followed by Chen et al.~\yrcite{Chen2021GraphNN} who leverage a Graph Convolutional Neural Network (GCNN) to learn the solution of steady-state laminar fluid flows around two-dimensional objects based on solutions of the Navier-Stokes equation. In contrast to these works, we do not optimize the grid space but rather study the generalizability across meshes with different shapes and the ability of the solver to learn superpositions. We thereby show that the network approximates a solution operator rather than simply a map of initial conditions to final solutions.

\textbf{Comparison to the \textit{Neural Operator Network}:}
Li et al.~\yrcite{Li2020NeuralOG,Li2021FourierNO}, use a data-driven approach to learn the solution of PDEs by training a model to map an initial condition to the final output. By doing so, this \textit{Neural Operator Network} learns the solution operator to the PDE without knowing its exact form.
Salvi et al.~\yrcite{Salvi2021NeuralSP} expand the Neural Operator Network by applying it to stochastic PDEs and incorporating noise into the driving force of the dynamics.
Brandstetter et al.~\yrcite{Brandstetter2022MessagePN} push the Neural Operator Method further by recognizing that numerical methods of classical solvers can be expressed as a special form of autoregressive message-passing protocols in GNNs. They apply this to various time-dependent PDEs for modeling fluid-like flow.

As the \textit{Neural Operator Network} by Li et al.~\yrcite{Li2020NeuralOG} ist most similar to our work, it warrants a deeper discussion: Li et al.~\yrcite{Li2020NeuralOG} explore generalization across meshes limited to different resolution leaving the question of shape generalization open, which is addressed in this paper. To successfully adapt to meshes of different resolution, Li et al. alter the mesh connectivity by connecting all nodes within a fixed radius before feeding the graph to the \textit{Neural Operator Network}. In our system, the GNN operates directly on FEM meshes where only directly adjacent nodes are connected. Therefore, generalization across resolution and shapes has to be learned by the operator without explicit clues. Within the \textit{Neural Operator Network} the important information about PDE inhomogeneities is provided via edge attributes to the GNN. In contrast to this, our operator directly operates on node features. The edge weights contain only relative distances between nodes. This formulation allows the usage of a broader class of GNN operators that do not need to incorporate extensive processing of edge attributes.

\section{Background}
\label{sec:background}

In this paper we consider a small but nevertheless important class of PDEs subject to boundary conditions. Let $u \in C^{2}(\mathbb{R})$ be a twice-differentiable function of $n$ variables over an open domain $\Omega \subset \mathbb{R}^n$. Then, the \textit{Poisson Equation} with \textit{Dirichlet} boundary conditions is given by
\begin{align}
    \nabla^2 u(x) &= f(x), \;&\forall x \in \Omega \nonumber, \\
    u(x) &= g(x), \;& \forall x \in \partial\Omega, \label{eq:Poisson_Dirichlet}
\end{align}
where $f$ and $g$ are sufficiently smooth functions defined on $\Omega$ and $\partial\Omega$, respectively, where $\partial\Omega$ denotes the boundary of the domain $\Omega$. The Poisson equation is often encountered in electrostatics or stationary wave (resonance) problems. Equation (\ref{eq:Poisson_Dirichlet}) is characterized by the absence of a time-component and the presence of a constraint given by $g$.

To solve equation~\eqref{eq:Poisson_Dirichlet} using an FEM solver, we cast it into its weak form using a test function $v$. We homogenize the boundary conditions with the transformation $y(x) = u(x) - g(x)$ and define $h(x) = f(x) - \nabla^2 g(x)$. In the last step, we used the continuation of $g$ into the domain $\Omega$. The weak form is then obtained using Green's first identity (also know as Stokes' theorem) and results in
\begin{align}
    a(y, v) = \hat{h}(v), \label{eq:weak_form}
\end{align}
where 
\begin{align}
    a(y, v) &= \int_\Omega \mathrm{d}x \sum_{i=1}^n\frac{\partial y}{\partial x_i}\frac{\partial v}{\partial x_i}, \nonumber \\
    \hat{h}(v) &= \int_\Omega \mathrm{d}x h(x)v(x).
\end{align}
The key insight of FEM is to choose the test function $v$ such that it is well behaved and can be represented as a sum over compact, local basis functions on $\Omega$, i.e. $v(x) = \sum_t \phi_t(x)$, where the basis elements $\phi_t$ are piece-wise linear functions or simplices, known as \textit{finite elements}. The solution $y(x)$ is then expressed as $y(x) = \sum_t y_t \phi_t(x)$ and $y_t$ is representing the value of the solution at a defined node of the simplex, i.e. a finite element. Using this basis expansion, equation~\eqref{eq:weak_form} reduces to the linear algebraic equation
\begin{align}
    \mathcal{A}\mathbf{y} = \mathbf{h}, \label{eq:FEM_Linear_equation}
\end{align}
where $\mathbf{y} = (y_t)_{t=1\dots T}$ and $\mathcal{A}$ can be calculated from the weak-form integral.

The choice of the finite element grid is still open, but it can be seen that the relations between the neighborhood of $y_t$ can be concisely expressed as edges between corresponding nodes in a graph structure $G$. Using this relationship, we use a GNN and graph convolutions to learn the solution operator of (\ref{eq:FEM_Linear_equation}).

\textbf{Poisson Equation for electrostatic and magnetostatic problems}: The \textit{Poisson Equation} can be used to explain many physical phenomena. In the following, we focus on experiments with electrostatics and magnetostatics simulations on two-dimensional meshes. The derivations we present are based on Langtangen et al.~\yrcite{langtangen2017solving}. The electric potential $U$ can be described with the following variant of the \textit{Poisson Equation}:
\begin{align}
-\nabla^{2} U = \frac{\rho}{\epsilon}, \label{eq:es}
\end{align}
where $\rho$ is the charge density and $\epsilon$ is the permittivity of the material. With the \textit{Dirichlet} boundary condition, we set $U$ at the boundary to a fixed value of zero. Note that we can still include PDEs with non-zero boundary conditions in this configuration: Our final operator would be applicable to non-zero boundary conditions by including the boundary to the solution domain and placing corresponding charges on those nodes. The solution can then be found by creating a new boundary outside of the solution domain, which can be discarded afterwards. After solving equation~\eqref{eq:es}, the electric field can be computed via the gradient of the potential: 
\begin{align}
-\nabla U = \mathbf{E}.
\end{align}
Let $x$ and $y$ be the spatial coordinates of the mesh. To simulate magnetostatic effects on a two-dimensional mesh, we consider electric currents along the $z$ axis that are orthogonal to the mesh. The magnetic vector potential $\mathbf{A}$ can be described as
\begin{align}
-\nabla^{2} \mathbf{A} = \mu\mathbf{I}, \label{eq:ms}
\end{align}
where $\mathbf{I}$ is the current density and $\mu$ is the permeability of the material. We consider the case $I_x = I_y = 0$. Our magnetostatics experiments can thus be viewed as sets of infinite wires that are orthogonal to the two-dimensional mesh. As the cross section of the wires looks the same irrespective of the $z$ coordinate, we find that $A_x$ and $A_y$ cannot depend on $z$. For symmetry reasons, the magnetic field is zero in $z$ direction and the interesting part of the vector potential can be reduced to a scalar value $A_z$. With the \textit{Dirichlet} boundary condition, we set $A_z$ at the boundary to a fixed value of zero. With a similar argument as above, our trained solution operator can generalize to non-zero boundary conditions as well. After solving equation~\eqref{eq:ms}, the magnetic field can be derived as the curl of $\mathbf{A}$, or in our case, simply:
\begin{align}
    B_x = \frac{\partial {A}_z}{\partial y} , B_y = -\frac{\partial {A}_z}{\partial x}.
\end{align}

\begin{figure*}
\centering
  \includegraphics[width=\linewidth]{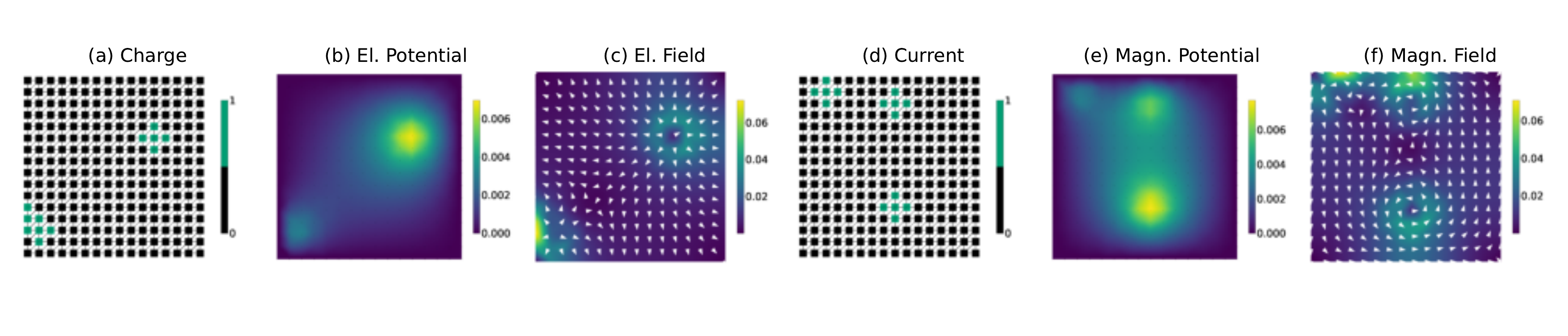}
  \vspace{-12mm}
\caption{Visualized input and output quantities for the square mesh. For the scalar fields (potentials) we plot the magnitude. For the vector fields, we visualize the magnitude of the field and overlay it with arrows depicting the field orientation.  (a) The charge input for the electrostatics problem. (b--c) Prediction targets in form of the potential and the electric field. (d) Current input for the magnetostatics problem. (e--f) Prediction targets in form of the magnetic vector potential (only $z$ component) and the magnetic vector field.}
\label{fig:features}
\end{figure*}

\begin{table}
\centering
\begin{tabular}{lcc}
               & PDE & Quantities  \\
               \toprule
Input   & es        & \makecell{Distance to border ($dx$, $dy$)\\ Charge ($\rho$)\\ Boundary condition} \\ \midrule
Output  & es        & \makecell{Electric potential ($U$)\\ Electric field ($E_x$, $E_y$)} \\
                 \toprule

Input   & ms    & \makecell{Distance to border ($dx$, $dy$)\\ Electric current ($I$)\\ Boundary condition} \\ \midrule

Output  & ms    & \makecell{Magnetic vector potential ($A_z$)\\ Magnetic field ($B_x$, $B_y$)}
\end{tabular}
\caption{Overview of the node attributes used as input and target for training of the GNNs. es: electrostratics, ms: magnetostatitcs.}
\label{tbl:in_out}
\end{table}

\section{GNN Solution Operator Experiments}
After discussing the variants of the Poisson equation that we aim to solve in section \ref{sec:background}, we now turn to the description of the concrete experiments and the data generation process. 

\subsection{Expressing PDE problems on graphs} 
The triangulation of the solution domain for the use of FEM solutions can naturally be expressed as a graph. Each node $p_i$ of the triangulation can be described by a binary feature $b_i \in\{0, 1\}$, denoting whether the point is a boundary point or interior point, and additional features $q_i$ describing the inhomogeneities of the PDE, i.e. $p_i=(b_i, q_i)$. We also use relative distances $\mathcal{D}_{ij}$ between connected nodes as edge attributes in the graph. See table \ref{tbl:in_out} for an overview of the node attributes. In figure \ref{fig:features} we show two examples of generated inhomogeneities and PDE solutions within our data.

The goal of our approach is to learn a neural network, such that
\begin{align}
    \mathbf{y} &= \mathbb{NN}_\theta[\mathbf{p}, \mathcal{D}], \textrm{s.t. } \mathcal{A}\mathbf{y} = \mathbf{h},
\end{align}
where $\mathbb{NN}_\theta$ is a neural network parametrized by $\theta$ that operates on the node features $\mathbf{p}$ and edge features $\mathcal{D}$ defined above. We train this network such that, given a triangulation mesh $\mathbf{x}$ and additional features, it predicts the solution values $\mathbf{y}$ of the underlying PDE. Note that an individual value $y_i$ can be a scalar or a vector depending on the underlying system described by the PDE. This corresponds to learning the solution operator in equation~\eqref{eq:FEM_Linear_equation}. Our final network is able to predict solutions for a variety of geometries with only a few graph convolution operations and faster than an FEM solver.  

We parameterize the model using a GNN where each node in the triangulation maps to a node in the GNN. In this way, our approach is extensible to modern FEM triangulation meshes that are typically not equidistant but dynamically refined as e.g. in~\cite{Pfaff2021LearningMS}. We train the network in a supervised fashion to predict the solution to a specific instance of the corresponding PDE that was obtained from a simulation via an FEM solver. Adapting the mesh resolution to underlying properties of the physical simulation would be a straightforward extension, which we defer to future work.

\begin{figure*}[t]
  \centering
  \subfigure[Visualization of the five mesh types without any mesh augmentation.]{
    \includegraphics[width=.6\textwidth]{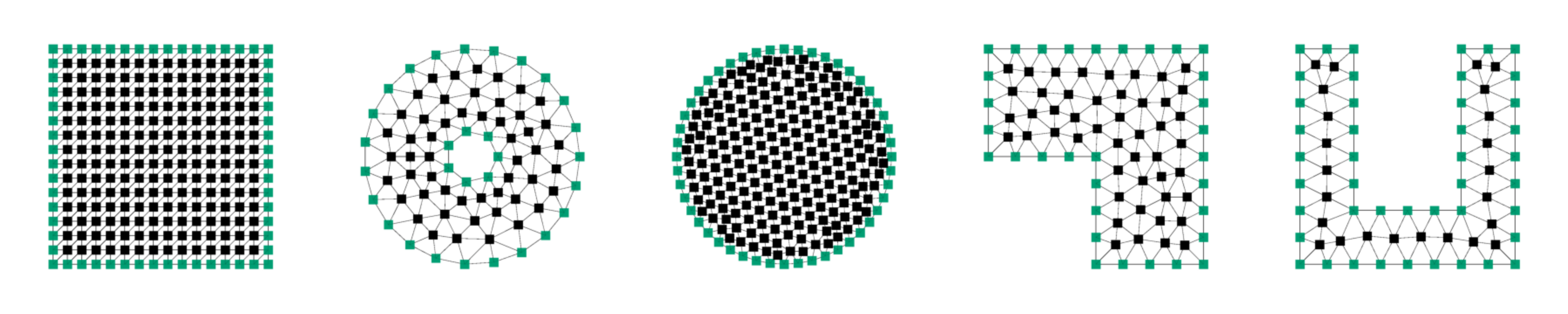}
    \label{fig:mashe_no_aug}
  }
  \subfigure[Examples of meshes with mesh augmentation. The U-mesh is not augmented.]{
    \includegraphics[width=.6\textwidth]{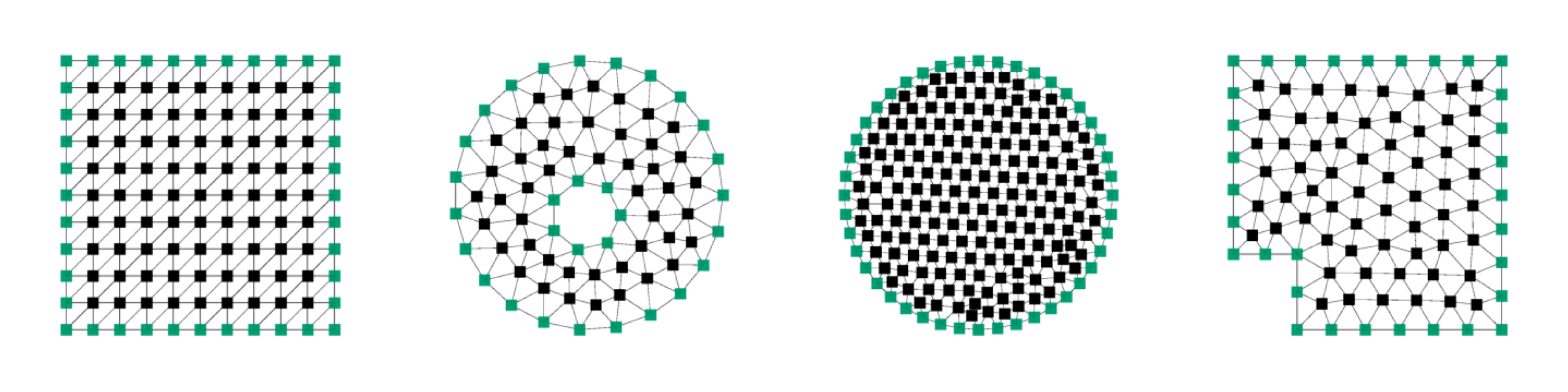}
    \label{fig:mashe_with_aug}
  }
  \caption{Examples of the different mesh geometries used for solving the PDEs and constructing the dataset. The augmentations are applied to (i) the disk mesh and square mesh, where we vary the node density (ii) the disk with a hole, where we vary the hole size and location (iii) the L-mesh, where we control the size of the cutout. The nodes where \textit{Dirichlet} boundary conditions are active are marked in green. Note that the different meshes also vary in resolution.}
  \label{fig:meshes}
\end{figure*}

\begin{table}[t]
\centering
\begin{tabular}{ccc}
   & Mesh augmentation & Number of charges / currents \\
   \toprule
   set1 & \xmark & 1--3 \\
   set2 & \cmark & 1--3 \\
   set3 & \xmark & 4--5
\end{tabular}
\caption{For each mesh, we generate sets of \num{2500} samples each: we generate data with and without mesh augmentation and also vary the number of positive charges or currents. For each set in the table we generate a total of \num{12500} samples, as we use 5 meshes. The data generation is applied equally for the electrostatics and magnetostatics simulations.}
\label{tbl:data_generation}
\end{table}

\subsection{Dataset Generation} \label{sec:ds_generation}

We generate datasets for several physical quantities over a variety of geometries using different simple simulations with the FEniCS library~\cite{scroggs2021construction}, a free FEM solver.

\textbf{Mesh geometries and mesh augmentations:} We use five different mesh geometries on which we solve the PDEs numerically. Our geometries are comparable to recent work~\cite{hsieh2019learning}: A square mesh, a disk with and without a hole, an L-shaped mesh and a U-shaped mesh. See figure~\ref{fig:mashe_no_aug} for a visualization of the meshes. We use normalized coordinates in the range [0,1] to define our meshes and thus do not report units for the coordinates. Our default square mesh has 256 evenly distributed nodes, the disk mesh has 252 nodes. To construct a disk with a hole, we use a disk with radius 0.5 and center (0.5, 0.5) and a circular cutout with center (0.5, 0.5) and radius 0.12. This yields 82 nodes for the disk with hole. For the L-mesh, we cut out one quarter of a square which yields 80 nodes. We construct a regular U-mesh with 68 nodes as a cutout from a unit square. The absolute positions of the nodes in the meshes are not relevant, as we do not use those during training. We use Delaunay triangulation to obtain the tesselations for all shapes.

In addition to the basic meshes, we induce variability to the data using \textit{mesh augmentations}. These augmentations act as a regularizer to the network similar to standard data augmentation techniques in other domains. We hypothesize that these augmentations are useful for training a network that is able to generalize to a variety of meshes with different geometries. The transforms consist of random variations of the mesh density of the disk and square shape, the location and radius of the hole in the disk as well as the size of the cutout in the L-mesh. A set of exemplary augmentations are depicted in figure \ref{fig:mashe_with_aug}. We do not apply augmentations to the U-mesh, as it is only used for testing generalization.

Specifically, for the square shape, we vary the number of nodes and thereby the resolution of the mesh ranging from 64 to 441 nodes. Similarly for the disk shape, we vary the number of nodes ranging from 63 to 411. For the disk with a hole, we vary the $x$ and $y$ coordinates of the cutout, both in the range of 0.35 to 0.65. We also vary the radius of the cutout from 0.05 to 0.25. For the L-mesh, we vary the width and height of the cutout rectangle both in the range of 0.2 to 0.8.

During training, we compare the proposed mesh augmentations with other standard approaches like node or edge dropout or dropout of embeddings (see section \ref{sec:training_details} for a more in-depth description).

\textbf{Physical problems:} Having described the mesh geometries underlying the dataset, we now turn to the physical PDEs we aim to solve on those meshes. We focused on simulating the two-dimensional electrostatics and magnetostatics problems described in section \ref{sec:background}. The magnetic vector potential as well as the electric potential are described by Poisson equations with current or charge inhomogeneities respectively. We also experiment with training the networks to predict other quantities like the magnetic or electric vector fields. 

In the case of electrostatics, we generated up to 3 circular charges for each instance. The charges are distributed randomly over the mesh and all have the same positive charge density. We consider the permittivity $\epsilon$ to be constant. Generalizing across different values for the permittivity and charge density should be possible with a minor modification of our dataset, but we chose to omit this for simplicity. We then solved the two-dimensional electrostatics problem numerically and saved the mesh with its metadata as well as the resulting field and potential solution as a single instance of the resulting dataset. For the magnetostatics experiments, we followed the same approach like above, but replaced the circular charges with an electric current and assume a constant permeability $\mu$. Again, generalizing across different values of the permeability has been omitted for simplicity.

For each of the meshes and PDEs, we created three groups of \num{2500} simulations: The first group was generated without any mesh augmentation. The second group uses mesh augmentation as described above. The third group includes more difficult simulations that are composed of simpler ones, such as examples with a higher number of charges or currents. The different settings are summarized in table \ref{tbl:data_generation}. We generated \num{7500} simulations for each mesh, i.e. \num{37500} simulations in total.


\subsection{Experiments}
\label{sec:experiments}

We create two different tasks to test the capabilities of our model with respect to its approximation quality and generalizibility. More specifically, we test (i) superposition, i.e. the networks' ability to generalize to a larger number of inhomogeneities by summing up the results seen during training. (ii) Shape generalization, i.e. the model's ability to generalize to previously unseen geometries. In both cases, we test the performance with and without mesh augmentation to demonstrate the importance of data-driven regularization. To evaluate these tasks, we need to split the data into training, validation and test sets accordingly. The splitting strategies apply equally to the magnetostatics and electrostatics experiments. We make the full dataset and experiment splits publicly available. 

We design a single experiment to test superposition: Considering all meshes, we test if the model can generalize from simple problems (e.g. electrostatics simulations with up to three charges) to more complex compositions (e.g. electrostatics simulations with up to five charges). For testing generalization across meshes of different shapes, we train on a subset of all meshes and use a single unseen mesh for testing. To make the shape generalization task more difficult, we chose to exclude the U-mesh, which exhibits less symmetries than others.

In total, we split the generated data (see table \ref{tbl:data_generation}) in two test sets and two training and validation sets:

\textbf{Training / validation set without mesh augmentation}: We select all generated data without mesh augmentation and 1--3 charges (set1 in table \ref{tbl:data_generation}) from the square mesh, L-mesh, circle and circle with a hole (all meshes except the U-mesh). We split this data into training and validation using a 80:20 split. There are a total of \num{2000} samples in the validation set and \num{8000} samples in the training set without mesh augmentation.

\textbf{Training / validation set with mesh augmentation}: We select all generated data with mesh augmentation and 1--3 charges (set2 in table \ref{tbl:data_generation}) from the square mesh, L-mesh, circle and circle with a hole (all meshes except the U-mesh). We split this data into training and validation using a 80:20 split. There are a total of \num{2000} samples in the validation set and \num{8000} samples in the training set with mesh augmentation.

\textbf{Shape generalization test set}: We select all generated data without mesh augmentation (set1 in table \ref{tbl:data_generation}) from the U-mesh. We have \num{2500} samples in this test set.

\textbf{Superposition test set}: We select all generated data without mesh augmentation and 4--5 charges (set3 in table \ref{tbl:data_generation}) from the square mesh, L-mesh, circle and circle with hole (all meshes except the U-mesh). To be of the same size like the shape generalization test set, we randomly sample 25\% of this data, such that there are \num{2500} samples in the superposition set.

The two tasks can be summarized as follows: For the \textit{superposition task} we train and validate two times, each time using one of the train and validation sets. We test each experiment on the superposition test set. For the \textit{shape generalization task} we train and validate two times, each time using one of the train and validation sets. We test each experiment on the shape generalization test set.

\subsection{GNN Architecture and Training Details.}
\label{sec:training_details}


Our model follows the encoder-processor-decoder architecture proposed in recent work on learning physical simulations with GNNs~\cite{Pfaff2021LearningMS}. The encoder and decoder are two-layer MLPs with 128 hidden units each and ReLU nonlinearities except for the output layer of the decoder, after which we do not apply any nonlinearity.  We use the spectral graph convolution operator introduced by Defferand et al.~\cite{defferrard2016convolutional} with 128-dimensional hidden features. We set the number of hops for the graph convolutions in the processor to $K=5$. With the same number of graph convolution layers, this allows capturing information across a broader receptive field in comparison to the \textit{Neural Operator Network}~\cite{Li2020NeuralOG}, which employs message passing only between neighboring nodes directly. We use 3 consecutive graph convolutions with a ReLU nonlinearity after each step. The model has a total of \num{280835} parameters. We implement the GNN based on PyTorch \cite{Pytorch2019} using the PyTorch Lightning framework \cite{PytorchLightning} as well as PyTorch Geometric \cite{PyTorchGeometric2019}.

We use the Adam optimizer \cite{kingma2014adam} with learning rate 1E-3 and betas 0.9 and 0.999. We use the mean squared error over all predicted quantities for training and reporting results. We use a batch size of 32 and train on an NVIDIA T4 with 16GB VRAM in a Google Kubernetes cluster. The node attributes, i.e. the input of the encoder, contain the charge (current) of the node, a flag indicating whether or not it is part of the boundary and the node's distance to the closest boundary node. We use relative distances to neighboring nodes as edge attributes. 

As a normalization strategy, all features are divided by their respective maximum absolute values, which are obtained from both accumulated train splits (see section \ref{sec:experiments}). The training data for all quantities thus lies in the range [-1,1]. We report results relative to the normalization constants without units. The network is tasked to predict the electric (magnetic) potential as well as the electric (magnetic) vector fields. 

We train each experiment for 150 epochs and validate after each epoch. For the final evaluation, we use the model parameters from the epoch with the lowest validation loss. We use the same random seed for all experiments.

We compare our proposed mesh augmentation with other regularization and augmentation techniques. We use dropout of node features before feeding them to the encoder with a probability of 0.2 (feature dropout), dropout of nodes in the graph with a probability of 0.1 and dropout of edges in the graph with a probability of 0.2. 

\subsection{Results}
\label{sec:results}

\begin{table}[t]
    \centering
    \begin{tabular}{lllll}
        PDE & Task & Mesh aug. & $\textrm{MSE}_{\textrm{pot}}$ & $\textrm{MSE}_{\textrm{field}}$ \\ \toprule
        es & shape & \xmark & $0.327\mathrm{E}{-3}$ & $2.723\mathrm{E}{-3}$ \\ 
        es & shape & \cmark & $\mathbf{0.006\mathrm{E}{-3}}$ & $\mathbf{1.821\mathrm{E}{-3}}$ \\ \midrule
        ms & shape & \xmark & $0.161\mathrm{E}{-3}$ & $2.640\mathrm{E}{-3}$ \\ 
        ms & shape & \cmark & $\mathbf{0.006\mathrm{E}{-3}}$ & $\mathbf{1.338\mathrm{E}{-3}}$ \\ 
        \toprule
        es & sup. & \xmark & $0.272\mathrm{E}{-3}$ & $0.602\mathrm{E}{-3}$ \\ 
        es & sup. & \cmark & $\mathbf{0.102\mathrm{E}{-3}}$ & $\mathbf{0.267\mathrm{E}{-3}}$ \\ \midrule
        ms & sup. & \xmark & $0.256\mathrm{E}{-3}$ & $0.434\mathrm{E}{-3}$ \\ 
        ms & sup. & \cmark & {$\mathbf{0.172E{-3}}$} & $\mathbf{0.276E{-3}}$ \\ 
    \end{tabular}
    \caption{Test results for electrostatics (es) and magnetostatics (ms) PDEs. Mesh augmentation (mesh aug.) denotes if the training/validation set including or excluding mesh augmentation is used. The best results are highlighted. The data for all experiments is normalized and thus results are multiples of normalization constants and reported without physical units. The top half shows the rest results in the shape generalization (shape) task: The shape generalization test set is used for these experiments. The bottom half shows the test results for the superposition (sup.) task: The superposition test set is used in these experiments.}
    \label{tbl:results_shape}
\end{table}

\begin{table}[t]
    \centering
    \begin{tabular}{llll}
        PDE & Regularization & $\textrm{MSE}_{\textrm{pot}}$ & $\textrm{MSE}_{\textrm{field}}$ \\ \toprule
        es & feature drop & $1.094\mathrm{E}{-3}$ & $4.104\mathrm{E}{-3}$ \\ 
        es & none           & $0.327\mathrm{E}{-3}$ & $2.723\mathrm{E}{-3}$ \\ 
        es & node drop      & $0.077\mathrm{E}{-3}$ & $3.077\mathrm{E}{-3}$ \\ 
    es & edge drop      & $0.041 \mathrm{E}{-3}$ & $2.826\mathrm{E}{-3}$ \\ 
        es & mesh augmentation & $\mathbf{0.006E{-3}}$ & $\mathbf{1.821E{-3}}$ \\ 
    \end{tabular}
    \caption{Study comparing mesh augmentation with various other regularization techniques for the electrostatics (es) PDE. The shape generalization test set is used in all experiments. The best results for each quantity and problem are highlighted.}
    \label{tbl:results_study}
\end{table}

\begin{figure}[!t]
\centering
\subfigure[Largest errors for Superposition.]{
  \includegraphics[trim={15cm 0 0 0},clip,width=.37\textwidth]{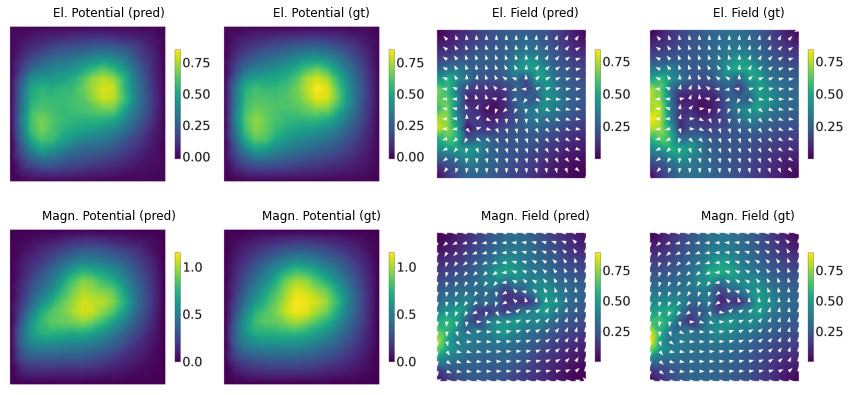}
}
\subfigure[Largest errors for shape generalization.]{
  \includegraphics[trim={15cm 0 0 0},clip,width=.37\textwidth]{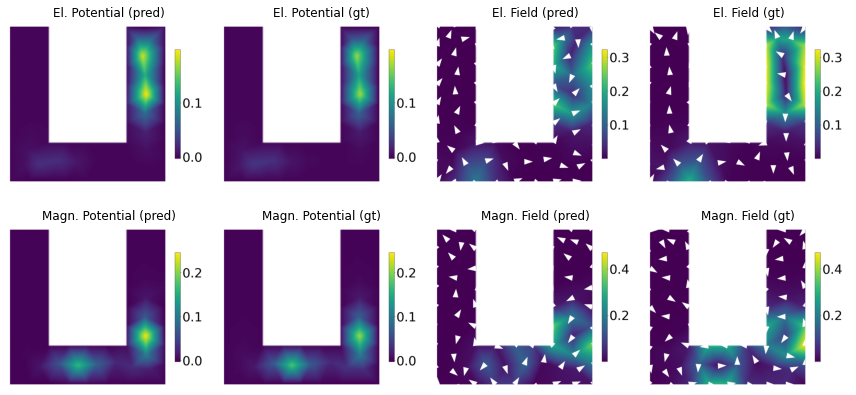}
}
\caption{Visualizations of the largest testing errors for our best performing models. We compare predictions of the model (pred) with ground truth data from the FEM simulation (gt). The background color depicts the magnitude of the respective field and the orientation is denoted by the arrows. The mean squared errors (vector field) for the shape experiments are $7.327\mathrm{E}{-3}$ (electrostatics) and $10.05\mathrm{E}{-3}$ (magnetostatics). The mean squared errors (vector field) for the superposition experiments are $1.013\mathrm{E}{-3}$ (electrostatics) and $1.154\mathrm{E}{-3}$ (magnetostatics).}
\label{fig:vis_results}
\end{figure}


We first show that the capability to solve PDEs on meshes with varying shapes is greatly enhanced by mesh augmentation during training. Table \ref{tbl:results_shape} summarizes the results for the shape generalization task for both the electrostatics and magnetostatics problems. Note that, due to the normalization of all quantities, we report results without units as described in section \ref{sec:training_details}. For this task, the model is presented with an unseen mesh (the U-mesh) during testing. Comparing the results for the electric (magnetic) potential, the version of the model with mesh augmentation outperforms the baseline without mesh augmentation by at least two orders of magnitude. For the derived quantities, i.e. the electric (magnetic) field, there is a smaller but still substantial improvement. Overall, mesh augmentation seems to be a crucial component for generalizing to unseen shapes.

Secondly, we show that mesh augmentation improves the performance for the superposition task as well. Table \ref{tbl:results_shape} summarizes the results for the superposition task for both the electrostatics and magnetostatics problems. For both the prediction of the potentials as well as the derived quantities, the electric and magnetic fields, we notice substantial improvements when using mesh augmentation. We therefore argue that mesh augmentation is a general technique to improve generalization for different kinds of tasks. More generally, diversity in the training data seems to improve the performance of the model with respect to a variety of tasks, which suggests that training on a diverse dataset with different meshes is a key ingredient for learning generalizable solution operators for static PDEs.

To investigate this further, we study other techniques to augment and regularize training and compare those with the proposed mesh augmentation. We exemplarily select the electrostatics problem to carry out this study. In general, those techniques that augment and vary the graph structure lead to improvements. Table \ref{tbl:results_study} shows the results for this experiment comparing mesh augmentation with random dropout of entire nodes or edges as well as dropout of node features and no augmentation at all. The dropout of features, which does not change or augment the graph structure, performs worse than no augmentation at all. Both dropout of edges and nodes change the mesh structure and thus contribute to better generalization. Mesh augmentation clearly outperforms all other augmentation techniques. This supports our argument that changes in the mesh structure during training are essential for learning generally applicable solution operators.

We further investigate the root cause of the comparatively high prediction error for the electric and magnetic fields. This behavior can be observed especially for the shape generalization task in table \ref{tbl:results_shape}. In figure~\ref{fig:vis_results}, we analyze the largest errors of our best-performing model for both the shape generalization and the superposition tasks. For the shape generalization task, we observe, that there is a mismatch in the prediction of the directions for both the magnetic and electric field. We also find that the model's predictions violate physical constraints like $\nabla\times\,\mathbf{E} = 0$ for the electric field.

\subsection{Runtime comparison}
\begin{figure}[t]
    \centering
    \includegraphics[width=\linewidth]{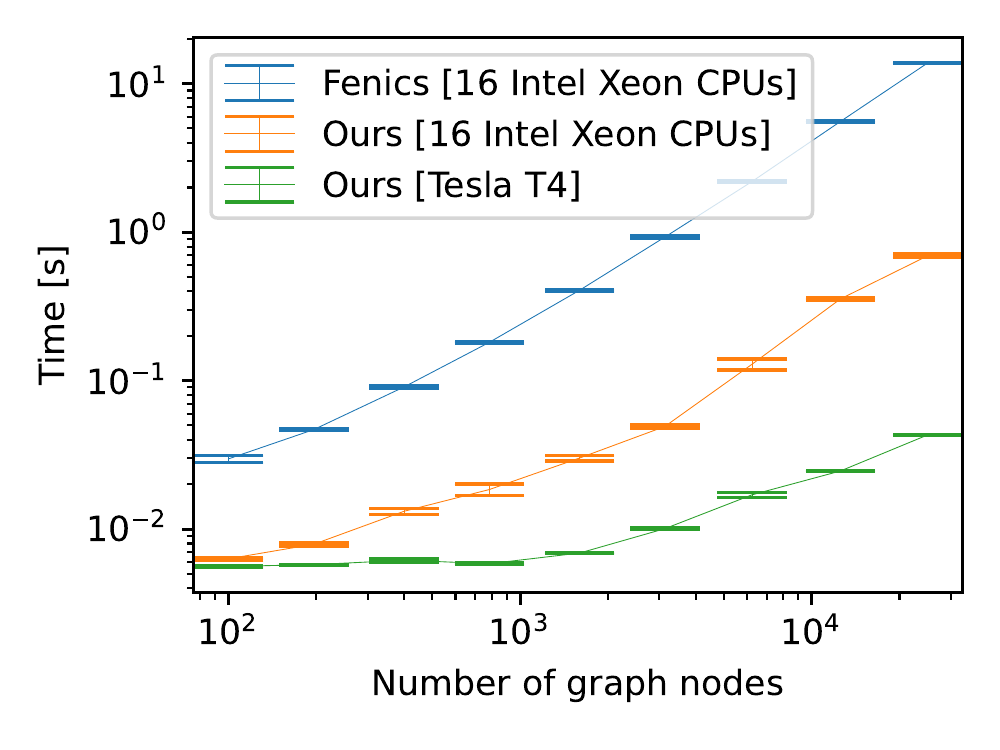}
    \caption{Runtime comparison of our GNN with the fenics FEM solver for predicting solutions on square meshes. We plot the time for solving a single example PDE versus the number of nodes in the mesh using a logarithmic scale. We compute each point on the curves as an average over 5 independent runs. The error bars depict the 95\% confidence interval.}
    \label{fig:wall_clock}
\end{figure}

To demonstrate the runtime benefits of our GNN operator, we compare the time to compute solutions for a fixed electrostatics PDE solved on a square mesh with varying resolution. We use 16 Intel(R) Xeon(R) CPU @ 2.80GHz cores for the study. A major advantage of our GNN operator is that it can be parallelized and executed on a GPU, while there is no GPU implementation available for the FEniCS library~\cite{scroggs2021construction}. We also evaluate the runtime using an NVIDIA T4 with 16GB VRAM for running inference with our trained GNN operator.

In figure~\ref{fig:wall_clock}, we compare the runtime of our trained operator with the iterative baseline. Our GPU-based implementation in PyTorch Geometric~\cite{PyTorchGeometric2019} is more than an order of magnitude faster than the baseline using an iterative FEM solver, while we already observe significant improvements for the CPU-based version due to parallelization on multiple CPU cores. Our results are similar to those of \cite{hsieh2019learning}, who observe faster execution using a CNN-based approximation of an iterative FEM solver.

\section{Summary and Conclusion}
To the best of our knowledge, we are the first to investigate the role of diversity in training data for learning accurate solutions operators in the case of static PDE problems. We show, that graph networks as approximators for static FEM simulations can generalize to unseen meshes with different shapes as well as an increased number of inhomogeneities with respect to charges or currents.

One of our key findings is that augmentation techniques on meshes are essential to enhance the ability of the neural operator to generalize. This suggests that graph neural networks can be used as universal solution operators for classes of PDEs, but only if they are trained carefully. Especially, diversity in the training data seems to be a key ingredient to enable robust generalization. 

In all our tasks, the electric or magnetic potential can be approximated with high accuracy. We therefore show that our system can solve the Poisson equation for a variety of different constraints. For the prediction of derived quantities like the electric or magnetic field, we observe comparatively large errors. As mentioned in section \ref{sec:results}, we observe that for these quantities the model violates physical constraints. 

Finally we argue that our GNN-based approach is flexible and able to generalize to a variety of different meshes, while it exhibits significant runtime advantages compared to iterative PDE solvers and heavily benefits from parallelization especially on GPUs.


We conclude that further research is needed to enable the learning of explicit or implicit physical constraints. As a main limitation of our work, it remains to be shown that more general classes of PDEs besides Poisson equations can be solved using our method. We aim at extending this study for a range of different problems and believe that our dataset should be used to verify the robustness for other neural operators in the future. 
The meshes that we use are generated using Delaunay triangulation with similar distances between all connected nodes. It remains to be shown that our approach generalizes to meshes with dynamically refined node densities.

\textbf{Contributions and Acknowledgments.}
WL lead the investigation, designed and implemented the experiments. SO supported with the experiment design, writing and delivered input on the physics problems. JSO helped in designing the study and experiment framework and oversaw the overall work. We would like to thank Maximilian Schambach for insightful discussions and feedback on the writing.

WL and JSO kindly acknowledge funding by the Federal Ministry for Economic Affairs and Climate Action (BMWK)
within the project ”KITE: KI-basierte Topologieoptimierung elektrischer Maschinen” (\#19I21034B). SO acknowledges funding by the DFG under SFB TR 185, Project Number 277625399.

\bibliographystyle{icml2022}
\bibliography{egbib}


\clearpage 
\section{Appendix}
\subsection{Elasticity data}
In an effort to create a more challenging dataset to foster future research, we create data for another PDE corresponding to gravity-induced deflection within a linear elasticity problem. The following derivations are based on Langtangen et al.~\yrcite{langtangen2017solving}. The initial equations, describing small elastic deformations, are
\begin{align}
    -\nabla\cdot\sigma &= f
    \\
    \sigma &= \lambda\text{Tr}\left(\epsilon\right)\text{I} + 2\mu\epsilon
    \\
    \epsilon &= \frac{1}{2}\left[\nabla u + \left(\nabla u\right)^\mathrm{T}\right].
\end{align}
Here, $f$ is the force acting on the body, $\lambda$ and $\mu$ are the Lamé coefficients, $\text{I}$ is the identity matrix, $\epsilon$ is the strain tensor, $\sigma$ is the stress tensor and $u$ is the displacement vector field. By combining the last two equations we obtain
\begin{align}
    \sigma\left(u\right)=\lambda(\nabla\cdot u)I+\mu(\nabla u+(\nabla u)^\mathrm{T}).
\end{align}
When solving these equations numerically, it is useful to work in a variational formulation. For this reason, we introduce an arbitrary test vector $v$ and integrate over the whole body to obtain the condition that must hold for any $v$:
\begin{align}
    a\left(u, v\right) = L\left(v\right)
\end{align}
where
\begin{align}
    a\left(u, v\right) &= \int_{\Omega}\sigma\left(u\right):\epsilon\left(v\right)\,dx,
    \\
    L\left(v\right) &= \int_{\Omega}f\cdot v\,dx
    + \int_{\partial\Omega_{T}}T\cdot v\,ds,
    \\
    \epsilon\left(v\right) &= \frac{1}{2}\left[\nabla v+(\nabla v)^T\right]
\end{align}
The colon operator represents the inner product between tensors (summed pairwise product of all elements) and $\partial\Omega_{T}$ is the part of the boundary where we prescribe the boundary condition $\sigma\cdot n=T$, with $n$ being the outward pointing surface normal vector.

\begin{figure}[h]
    \centering
    \includegraphics[width=\linewidth]{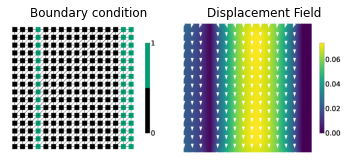}
    \caption{Elasticity input boundary conditions and displacement field.}
    \label{fig:in_out_elasticity}
\end{figure}

The data generation process for the elasticity problem is exactly the same as described in section \ref{sec:ds_generation}. See figure \ref{fig:in_out_elasticity} and table \ref{tbl:in_out_elasticity} for an overview of the input and output quantities we use for the linear elasticity data. For set1 and set2 in table \ref{tbl:data_generation}, we use 1--3 fixed vertical lines (via boundary conditions). For set3, we use an increased number of 4--5 fixed vertical lines to test superposition. For future reference, we will also publish training/validation/test splits like described in section \ref{sec:experiments} for this part of the dataset, although we did not conduct any experiments yet.

\begin{table}[h]
\centering
\begin{tabular}{lccc}
               & Linear Elasticity  \\
               \toprule
Input          & \makecell{Distance to border ($dx$, $dy$) \\ Distance to boundary condition ($dx$, $dy$) \\ Boundary condition} \\
               \midrule
Output         & \makecell{Displacement field ($u_x$, $u_y$)} \\
\end{tabular}
\caption{Overview of the node attributes used as input and target for the linear elasticity problem.}
\label{tbl:in_out_elasticity}
\end{table}

\subsection{Reproducibility study}
As the experiments were run only once for the paper, they were hard to reproduce using the original code. We provide additional results replicating the results of the paper by running all experiments using 5 different random seeds in table \ref{tbl:add_results_shape} and \ref{tbl:add_results_study}. In table \ref{tbl:add_complete} we provide detailed results for all combinations of pde, task and regularization using 5 different seeds. We provide both the mean scores and standard deviations. For all additional experiments we changed only the random seed. All other training details remain exactly as described in section \ref{sec:training_details}.

\begin{table}[t]
    \centering
    \begin{tabular}{lllll}
        PDE & Task & Mesh aug. & $\textrm{MSE}_{\textrm{pot}}$ & $\textrm{MSE}_{\textrm{field}}$ \\ \toprule
es & shape & \xmark & $0.165\mathrm{E}{-3}$ & $2.768\mathrm{E}{-3}$ \\
es & shape & \cmark & $\mathbf{0.023\mathrm{E}{-3}}$ & $\mathbf{2.224\mathrm{E}{-3}}$ \\
\midrule
ms & shape & \xmark & $0.077\mathrm{E}{-3}$ & $2.402\mathrm{E}{-3}$ \\
ms & shape & \cmark & $\mathbf{0.012\mathrm{E}{-3}}$ & $\mathbf{1.857\mathrm{E}{-3}}$ \\
\toprule
es & sup. & \xmark & $0.393\mathrm{E}{-3}$ & $0.872\mathrm{E}{-3}$ \\
es & sup. & \cmark & $\mathbf{0.209\mathrm{E}{-3}}$ & $\mathbf{0.482\mathrm{E}{-3}}$ \\
\midrule
ms & sup. & \xmark & $0.292\mathrm{E}{-3}$ & $0.600\mathrm{E}{-3}$ \\
ms & sup. & \cmark & $\mathbf{0.228\mathrm{E}{-3}}$ & $\mathbf{0.477\mathrm{E}{-3}}$ \\
    \end{tabular}
    \caption{Additional results extending table \ref{tbl:results_shape}. Results are means of test scores averaged over 5 random seeds.}
    \label{tbl:add_results_shape}
\end{table}

\begin{table}[t]
    \centering
    \begin{tabular}{llll}
        PDE & Regularization & $\textrm{MSE}_{\textrm{pot}}$ & $\textrm{MSE}_{\textrm{field}}$ \\ \toprule
es & feature drop & $1.575\mathrm{E}{-3}$ & $4.187\mathrm{E}{-3}$ \\
es & none & $0.165\mathrm{E}{-3}$ & $2.768\mathrm{E}{-3}$ \\
es & node drop & $0.067\mathrm{E}{-3}$ & $2.599\mathrm{E}{-3}$ \\
es & edge drop & $0.081\mathrm{E}{-3}$ & $2.941\mathrm{E}{-3}$ \\
es & mesh augmentation & $\mathbf{0.023\mathrm{E}{-3}}$ & $\mathbf{2.224\mathrm{E}{-3}}$ \\
 \end{tabular}
    \caption{Additional results extending table \ref{tbl:results_study}. Results are means of test scores averaged over 5 random seeds.}
    \label{tbl:add_results_study}
\end{table}

\begin{table*}[t]
\begin{center}\begin{tabular}{ c c c c c }
PDE & Task & Regularization & $\textrm{MSE}_{\textrm{pot}}$ & $\textrm{MSE}_{\textrm{field}}$ \\
\toprule
es & shape & none & $1.65E-04\,(6.44E-05)$ & $2.77E-03\,(1.27E-04)$ \\
es & shape & feature drop & $1.58E-03\,(9.28E-04)$ & $4.19E-03\,(1.05E-03)$ \\
es & shape & node drop & $6.69E-05\,(2.20E-05)$ & $2.60E-03\,(2.18E-04)$ \\
es & shape & edge drop & $8.09E-05\,(2.60E-05)$ & $2.94E-03\,(1.33E-04)$ \\
es & shape & mesh augmentation & $\mathbf{2.31E-05\,(1.01E-05)}$ & $\mathbf{2.22E-03\,(9.05E-05)}$ \\
\midrule
es & sup & none & $3.93E-04\,(2.66E-05)$ & $8.72E-04\,(3.69E-05)$ \\
es & sup & feature drop & $2.29E-03\,(2.96E-04)$ & $1.77E-03\,(7.70E-05)$ \\
es & sup & node drop & $1.60E-03\,(8.85E-05)$ & $1.21E-03\,(3.55E-05)$ \\
es & sup & edge drop & $8.96E-04\,(4.40E-05)$ & $8.89E-04\,(3.23E-05)$ \\
es & sup & mesh augmentation & $\mathbf{2.09E-04\,(5.50E-05)}$ & $\mathbf{4.82E-04\,(2.51E-05)}$ \\
\midrule
ms & shape & none & $7.70E-05\,(1.62E-05)$ & $2.40E-03\,(9.56E-05)$ \\
ms & shape & feature drop & $1.42E-03\,(8.80E-04)$ & $3.94E-03\,(4.33E-04)$ \\
ms & shape & node drop & $5.28E-05\,(1.60E-05)$ & $2.19E-03\,(1.48E-04)$ \\
ms & shape & edge drop & $8.67E-05\,(3.41E-05)$ & $2.67E-03\,(2.83E-04)$ \\
ms & shape & mesh augmentation & $\mathbf{1.24E-05\,(3.77E-06)}$ & $\mathbf{1.86E-03\,(1.23E-04)}$ \\
\midrule
ms & sup & none & $2.92E-04\,(5.09E-05)$ & $6.00E-04\,(3.38E-05)$ \\
ms & sup & feature drop & $2.02E-03\,(3.34E-04)$ & $1.31E-03\,(4.87E-05)$ \\
ms & sup & node drop & $1.43E-03\,(1.13E-04)$ & $8.38E-04\,(3.08E-05)$ \\
ms & sup & edge drop & $7.39E-04\,(1.39E-04)$ & $6.26E-04\,(3.20E-05)$ \\
ms & sup & mesh augmentation & $\mathbf{2.28E-04\,(3.11E-05)}$ & $\mathbf{4.77E-04\,(2.10E-05)}$ \\
\midrule
\end{tabular}\end{center}
    \caption{Additional results for all combinations of pde, task and regularization. Results are averaged over 5 random seeds. We provide the standard deviation in brackets.}
    \label{tbl:add_complete}
\end{table*}

\end{document}